# Creativity Reconsidered: Generative AI and the Problem of Intentional Agency

James S. Pearson (University of Amsterdam; University of Lisbon)
Matthew J. Dennis (TU Eindhoven)
Marc Cheong (University of Melbourne)

## Abstract

Many theorists maintain that conscious intentional agency is a necessary condition of creativity. We argue that this requirement, which we call the *Intentional Agency Condition* (IAC), should be abandoned. We motivate this by highlighting the problems this criterion encounters in the face of recent advances in generative AI, which is ostensibly creative despite being incapable of intentional agency. We present two corpus analyses to illustrate the rapidly increasing tendency of people to predicate creativity to generative AI. In response to this predicament, theorists of creativity have proposed a range of conflicting solutions, which we critically evaluate. We find that none of these satisfyingly resolves the initial predicament, and we therefore propose a novel approach. Our claim is that ascriptions of creativity are dependent on what we call *creative ability*. This solution explains why intentional agency is important for judgements of creativity, without being a necessary condition. Our approach thereby accommodates AI creativity without dismissing the intuition that perceived intentions are of key importance for ascriptions of creativity.

## 1. Introduction

In recent decades it has become common for theorists of creativity to posit intentional agency as a necessary condition of creativity (Gaut 2010; Livingston 2018; Runco 2023; Stokes and Paul 2021). 'Intentional agency' is in this context generally taken to mean acting in accordance with *consciously* represented and desired goals. According to this view, for an agent to qualify as 'creative,' they need to be capable of producing novel, valuable and surprising ideas and artefacts by means of purposeful, self-aware action. We call this requirement the *Intentional Agency Condition* (IAC). The principal intuition that is used to justify the IAC is that when someone produces something novel and valuable by accident, we would not be inclined to consider that person to have exhibited creativity (Gaut 2010). However, following others (e.g., Helliwell 2026b; Langland-Hassan 2026; McClelland 2026), we argue that the IAC has become saliently problematic in light of recent advances in generative AI, which is ostensibly creative without satisfying the IAC – i.e., generative AI is able to produce novel and valuable outputs almost independently, in a manner that departs from the intentions of programmers and prompters alike (Boden 2019; Nannicelli 2025: 1).[1] Further, there is a growing tendency to ascribe creativity to AI, a shift that is by no means confined to theoretical discussions. Indeed, we present two corpus analyses – one from Google Ngram Viewer and another from the News on the Web (NOW) corpus – that indicate just how comfortable Anglophone authors and journalists have become predicating creativity to generative AI. The pressing question to which this gives rise is as follows: How do we reconcile this tendency with the arguments that are typically invoked to justify the IAC? We refer to this problem as the *IAC predicament*. In response to this, we contend that there are good reasons to abandon the IAC, and show how doing so has the added benefit of enabling us to accommodate AI creativity.

Theorists have sought to address the IAC predicament in a variety of competing ways, and their responses can be broadly divided into two opposed groups: a *denialist* camp on the on side and an *accommodationist* camp on the other. Those belonging to the former generally seek to resolve the issue by doubling down on the IAC and denying the possibility of AI creativity. As such, many of these theorists insist that generative AI systems are only capable of "*pseudo*-creativity" (Runco 2023: 1) or "apparent" creativity (Boden 2014: 224), but not *genuine* creativity (see also Brainard 2025; Gaut 2010; Hoel 2022; Stokes and Paul 2021). Others, though, argue that our theoretical definition of creativity either already can, or with some straightforward revisions ought to be able to, accommodate AI creativity (Amabile 2020;

[1] By generative AI, we are especially referring to large language models (LLMs) and text-to-image or text-to-sound diffusion models such as Dall-E, Stable Diffusion, Midjourney, and Suno.

Helliwell 2026b; McClelland 2026; Langland-Hassan 2026). To the best of our knowledge, the key responses to the IAC predicament are yet to be collated and critically assessed. Far from being an abstract exercise in definitional hair splitting, we argue that these conflicting responses have significantly different practical ramifications.

Although we find existing responses to the IAC predicament to be either unpersuasive or incomplete, we nonetheless find a serviceable starting point in the current literature – especially in Peter Langland-Hassan's (2026) notion of creativity as a form of 'tuning' between the generative agent and its creative output. Building on this view, we argue that the most effective solution to the IAC predicament – i.e., the solution that best explains the growing tendency to predicate creativity to AI, while simultaneously accounting for the intuitions that originally motivated the IAC – is one that conceives creativity as an *ability*. This ability is then best understood as being grounded in a certain form of functional organization – one that is counterfactually capable of producing novel, valuable and surprising ideas and artefacts. Our claim is that when we ascribe creativity to an agent, we are inferring that the agent is functionally organized in precisely this kind of way.[2] Since we can never be *certain* that any agent possesses this type of creative functional organization, the question is what warrants us to *infer* this type of internal architecture. In addressing this question, we argue that perceived intentions are an important source of such warrant, especially in the case of humans, but they are certainly not the *only* source, nor are they a *necessary* one.

## 2. Why Demand Intentional Agency?

According to what has become known as the 'standard definition', creativity requires a) originality (or 'novelty') and b) effectiveness (also expressed as 'appropriateness', 'utility,' or 'value') (Runco and Jaeger 2012). However, many commentators, posit further necessary criteria. One of the most common of these being *surprise*. Thus, Boden (2004: 1) argues that "Creativity is the ability to come up with ideas or artefacts that are new, surprising and valuable". Various other additional conditions have been proposed – for instance, that creative objects must issue from 'curiosity' (Brainard 2025), or 'imagination' (Kind 2022), or 'flair' (which combines a cluster of different capacities) (Gaut 2010). But one recurrent supplement to the standard definition is particularly relevant given recent developments in generative AI – namely, that creativity requires intentional agency (i.e., the IAC). The foremost proponent of this criterion is Gaut (2010), though it has also been defended by a significant number of other theorists (e.g., Brainard 2025; Kieran 2014; Livingston 2018; Paul and Stokes 2018; Stokes 2011; Stokes and Paul 2021).

[2] Throughout this paper we will be using the term 'agent' to refer to any goal-directed system (following others, including Floridi 2025 and Dung 2025).

Theorists typically add the IAC to the standard definition to defuse a range of counterexamples. One of the strongest of these is offered by Gaut (2010: 1040), who asks

> [s]uppose that in walking through a studio you accidentally knock over a set of paints, which spill onto a canvas, and they happen to combine to produce a beautiful and original painting. You would not be creative, since the painting was purely the product of luck.

According to Gaut, the individual who knocked over the set of paints cannot legitimately be called 'creative' because they did not consciously *intend* to bring about the resulting painting.

For Gaut, the same applies in the case of generative processes that we witness in nature. Although a tree may produce an "elegant and distinctive canopy of leaves", it "is not acting at all, since it lacks desires, beliefs and other intentional states; so *a fortiori* it cannot be acting creatively" (Gaut 2018: 129; Gaut 2010: 1040). Others also exclude non-agential natural phenomena from the domain of genuine creativity (e.g., Livingston 2018: 116; Paul and Stokes 2018). Paul and Stokes (2018: 200–4) do so by appealing to our intuitions about linguistic practice. As they see it (202), if somebody beholding a sunset were to state, "This sunset is creative", we would take them "to have spoken nonsense" because the process that gave rise to the sunset "does not involve agency in any relevant way." They also defend the IAC with an argument based on justificatory practice, arguing that people typically appeal to certain processes (especially intentional agency) when asked to justify their ascriptions of creativity (Paul and Stokes 2018: 200–201).

These arguments often rest on either explicit or implicit claims about the intuitive inappropriateness of ascribing creativity to agents that have not exhibited any intentional agency in generating a given novel and valuable object. What is notable about the standard definition is that it allows for *any* agents that generate original and novel objects to be considered creative, regardless of the generative processes that yielded these products. When the IAC is added to the standard definition, however, creativity becomes markedly anthropocentric.

Some have sought to explain *why* it is intuitively incorrect to ascribe creativity to subjects that have not exercised intentional agency in generating a given novel and original idea or artefact. The dominant explanation hinges on the observation that ascriptions of creativity are *approbative* (Kieran 2014; Paul and Stokes 2018; Stokes 2011; 2014). Paul and Stokes (2018: 197) thus argue that to call someone 'creative' is to express praise towards that person, and "praise is not appropriately given to subjects who lack responsibility for their actions." As such, intentional agency appears to be a necessary condition of creativity.

What we wish to draw attention to here is that justifications for the IAC are predominantly based on theorists' intuitions about how the language of creativity is ordinarily employed. As Currie and Turner (2023: 119) observe, "Connections between creativity and agency [i.e., conscious, intentional agency] are typically driven by intuitions." In the following section, however, we are going to see that advances in generative AI are putting these intuitions under increasing pressure.

## 3. The IAC Predicament

Generative AI problematises the IAC because it routinely yields outputs that laypeople and experts alike judge to be creative. It consistently produces new, valuable, and surprising products in design, painting, architecture, poetry, etc., and does so in a manner tailored to users' prompts. These results are neither accidents, nor are they naturally occurring, automatically generated phenomena.[3] Moreover, AI has been shown to outperform most humans on a range of standardized creative tests (Guzik et al. 2023; Haase and Hanel 2023). Reflecting on the outputs of generative AI, multiple theorists of creativity judge AI to be genuinely creative. Ethan Mollick (2023), for instance, writes that "AI can generate creative ideas in real-life, practical situations," and Hubert et al. (2023) go even further, claiming that "AI is more creative than humans". And certain philosophers of creativity argue that AI is creative even while they very explicitly acknowledge that it lacks a capacity for intentional agency (Langland-Hassan 2026; McClelland 2026).[4] Peter Langland-Hassan (2026: 545), for example, draws this conclusion after considering the "strong *prima facie* evidence that contemporary generative AI is creative".

While Langland-Hassan is referring to everyday forms of creativity, others are confident that these systems will eventually be able to engage in the higher kinds of creativity that we associate with major scientific breakthroughs, or artistic genius – that is, what is often called ‘big-C’ creativity.[5] Some go so far as to claim that this ambitious goal has already been achieved. For instance, when DeepMind’s AlphaGo beat Lee Sedol, the South Korean go champion, commentators framed AlphaGo’s performance in precisely these terms (Halina 2021; Metz 2016).

---

[3] Though computational systems do also sometimes enable chance discoveries (Green 2004).
[4] James Kaufman also subscribes to this view (Marchant 2025).

This tendency to ascribe creativity to AI is also more generally growing among Anglophone authors and journalists. This trend is evident in two corpus analyses that we recently performed, one using Google Books Ngram corpus (Google 2024), and another using the News on the Web (NOW) corpus (see figures 1 and 2). In our analysis of the Google Books Ngram corpus, we tracked the use of language surrounding AI in published books from 1900 to 2022 (inclusive).[6] We selected the following bigrams to compare how the adjective ‘creative’ was predicated to AI in contrast to a range of subjects that are more conventionally considered to be capable of satisfying the IAC: “creative AI”, “creative musician”, “creative painter”, and “creative engineer”. Figure 1 shows a sharp upward spike for the bigram “creative AI” since *circa* 2017, coinciding with that year’s so-called ‘AI summer’ or ‘AI boom’ (Lohr 2017).

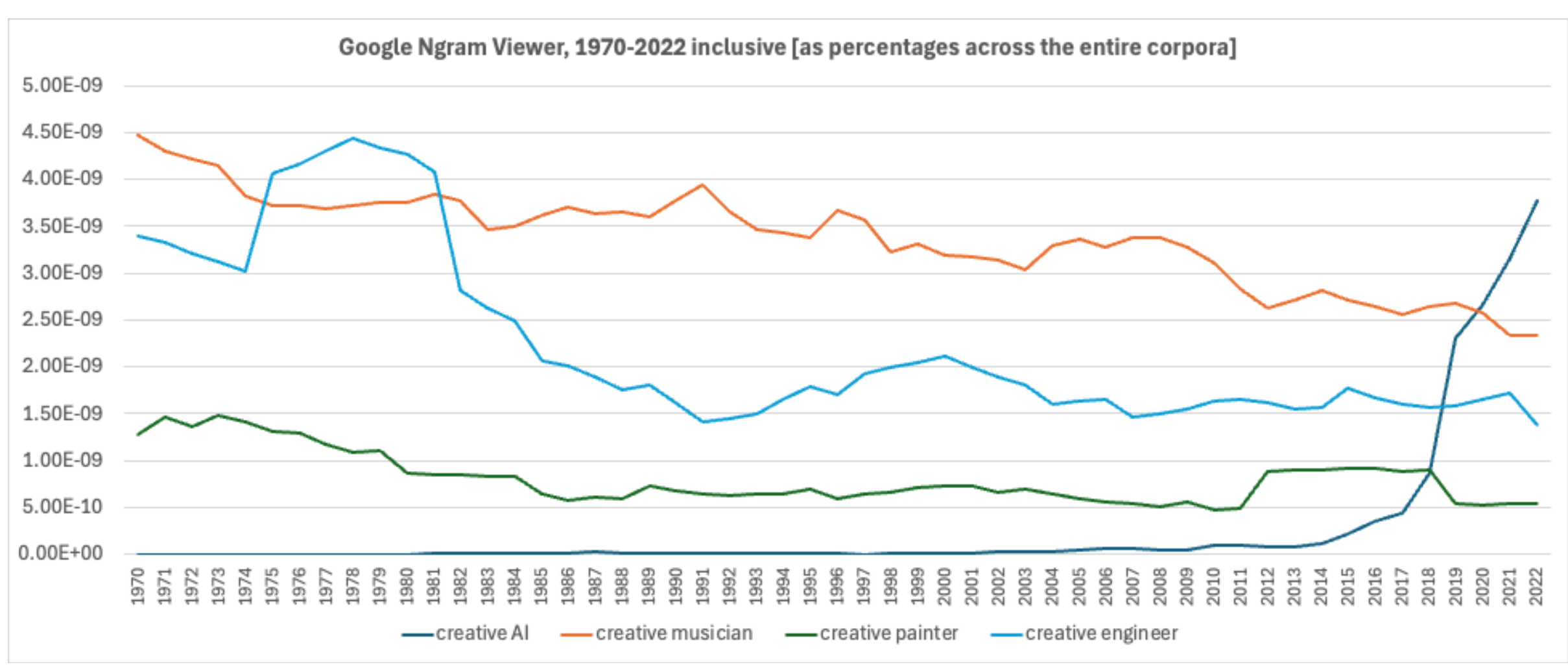

**Figure 1. Google Ngram Viewer trend data for {*creative AI, creative musician, creative painter, creative engineer*}, 1970–2022 inclusive. The score returned by Google Ngram Viewer is expressed as a percentage of all available bigrams in Google’s corpus, up to and including the year 2022 (at time of writing). Adapted from raw data provided by Google Books (2024); search conducted 3 Mar 2026 at URL = https://books.google.com/ngrams/graph?content=creative+AI%2Ccreative+musician%2Ccreative+painter%2Ccreative+engineer&year_start=1970&year_end=2022&corpus=en&smoothing=3&case_insensitive=false**

[5] See the discussion of Hiroaki Kitano in Hutson (2023). See also Colton (2019).

[6] An n-gram is a sequence of *n* contiguous words (hence, a bigram is simply a 2-gram, or two contiguous words). Ngram Viewer is a Google tool for analyzing n-gram frequencies in published books since *circa* 1500 CE (see Michel et al. 2011 for a full technical explanation).

We also analysed discussion of AI in news media, using the source corpus News on the Web, or NOW (Figure 2). This corpus houses "23.5 billion words of data from web-based newspapers and magazines from 2010 to the present time" (Davies, 2016–). NOW allows searches for the same bigrams, to provide us with annual trends of textual utterances and phrases, in absolute frequencies, over time.

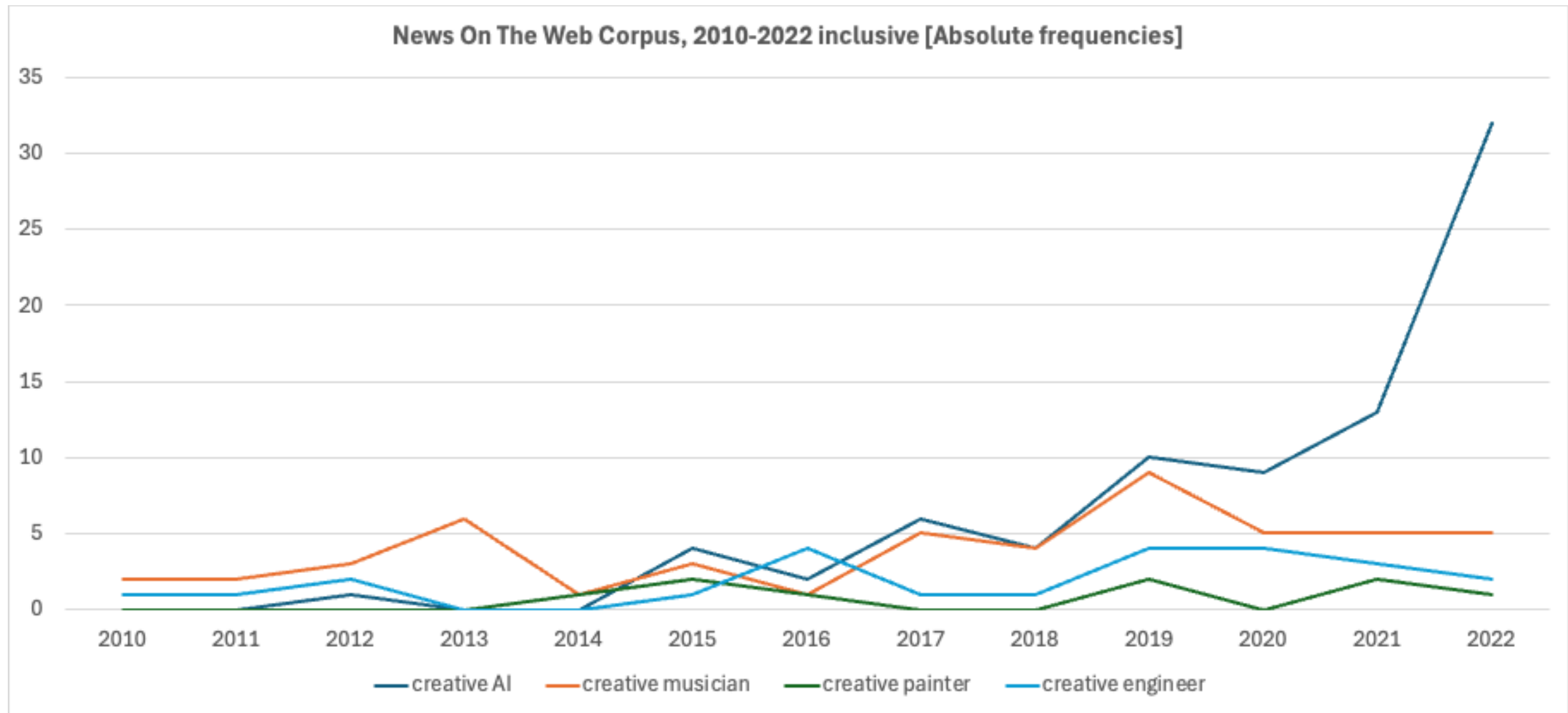

**Figure 2. News on the Web trend data for {*creative AI, creative musician, creative painter, creative engineer*}, 2010–2022 (inclusive). Adapted from raw data provided by Davies (2016–); search conducted 3 Mar 2026 at URL = https://www.english-corpora.org/now/.**

Trends in both the Google Books Ngram and NOW corpora show a recent uptick in ascriptions of the adjective 'creative' to AI. We do not present these empirical analyses as conclusive proof that people consider AI to be creative. Individual occurrences of the bigram "creative AI" might be part of statements that are denying the possibility of creative AI, or part of neutral commentaries on the current discourse on artificial intelligence. And where this usage is affirmative, it might not reflect people's *considered* intuitions. However, we believe that these objections are not sufficient to refute the hypothesis that authors and journalists are increasingly comfortable describing AI as creative. Furthermore, the reality of the IAC predicament does not stand or fall with these corpus analyses, which we present here to give additional credence to the tendency that we have already identified within the specialized field of creativity research.

It strikes us as reasonable to require that theoretical accounts of creativity broadly cohere with current use of the vocabulary of creativity, especially within expert discussions of creativity, and the writings of authors and journalists more generally. The fact that we find creativity being ascribed to AI in all these spheres of discourse does not *refute* intuition-driven defences of the IAC, but it does put heavy pressure on their claim to reflect stable or widely shared linguistic dispositions. That said, the appeals to intuition cited in support of the IAC may have been quite plausible and unproblematic in their time, since many of them were made before "creative AI" became a relatively common turn of phrase (i.e., before 2020). Thus, although these appeals to intuition may once have been quite valid, they have since fallen out of joint with both ordinary and expert language use.

To reiterate, the recent uptick in ascriptions of creativity to AI disrupts the IAC because it suggests that expert and non-expert anglophone authors are intuitively inclined to describe AI as creative despite it being ostensibly incapable of satisfying the IAC. AI systems are to all appearances unable to mentally represent (even approximately) or desire their target outputs prior to producing them. In the words of Salles et al. (2020: 694): "intentional actions are explained in terms of mental states (e.g., beliefs, desires, attitudes) that are the causal antecedents of an agent's behavior. No such mental states could be attributed, plausibly, to current AI, which means that attributing this kind of agency to AI systems is a mistake …".

This gives rise to the IAC predicament: many of the intuitions that originally motivated the IAC seem philosophically plausible – especially the claim that 'accidental creativity' is incoherent – and yet they simultaneously appear to be increasingly untenable in light of the growing number of theorists, authors and journalists ascribing creativity to AI. This issues in a dilemma: do we reject the IAC? Or deny AI creativity?

## 4. Possible Responses to the IAC Predicament

The candidate solutions to the IAC predicament can broadly be divided into two groups. On the one side, denialist approaches simply reject the possibility of creative AI and frame ascriptions of creativity to AI as erroneous. On the opposite side, accommodationist approaches maintain that AI creativity *can* be accommodated, either by our existing conception of creativity, or by some revised version thereof. We will consider the strengths and weaknesses of each of these in turn before outlining our own accommodationist approach.

### 4.1 Denialist Approaches

Theorists who deny that AI systems could ever be considered genuinely creative typically do so on the basis of what is commonly called a 'process-first' approach to creativity. This means that they insist that in order to legitimately ascribe creativity to an agent, we must first ensure that the agent has generated novel, valuable and surprising outputs by means of specific *processes*. Those who adhere to the process-first approach refer to a range of specific cognitive processes they take to be essential preconditions of creativity, and some explicitly invoke these to exclude AI from the domain of genuine creativity – for example, imagination (Gaut 2003; Kelly 2019; Kind 2022); emotion (Boden 2014; Hoel 2022); understanding, skill and flair (Gaut 2010); and curiosity (Brainard 2025). However, intentional agency stands out as the most core process requirement posited by theorists (as noted by Paul and Stokes 2018: 197), and is regularly cited as grounds to reject the notion of AI creativity (Gaut 2010; Runco 2023). So, although theorists might deny AI creativity by invoking various process conditions, we will exclusively focus on denialist approaches that appeal to the IAC.

*Overlooking the IAC*. Those advancing this particular denialist response accept that AI agents can produce outputs that resemble the fruit of genuine creativity; however, according to this view, people *mistakenly* ascribe creativity to the source AI agent because they neglect to verify whether the IAC has actually been satisfied (Hoel 2022; Runco 2023). This cluster of denialists claim that upon closer scrutiny AI agents are therefore merely "*pseudo*-creative" (Runco 2023). Although some members of this camp suggest that AI systems might in the future conceivably *become* conscious and capable of intentional agency (e.g., du Sautoy 2019, Chap. 16), the general view tends to be that AI is constitutionally incapable of satisfying the IAC.

The chief issue for this process-first denialist approach is that it requires that one adopt an essentialist and heavily prescriptivist stance.[7] Given that psychologists and philosophers of creativity are generally more interested in *human* creativity, the preponderance of anthropocentric definitions within these fields should perhaps come as no surprise. However, the problem with these definitions is that they exclude the common and persistent tendency of experts and ordinary language users to ascribe creativity to non-human entities, such as trees and evolutionary processes. Experts even attribute creativity to such phenomena after explicitly acknowledging the fact that they lack a capacity for conscious intentional agency (Arnheim 2001; Currie and Turner 2023). And as we have just seen, this is similarly the case with generative AI. Those who insist on the IAC end up positing a conception of creativity that fails to connect with people's actual use of the term. By contrast, it would be ideal if we had a theory of creativity that could *explain* expert and lay usage instead of simply branding large swathes of it erroneous.

[7] Though N.B. that Boden (2014, 226) is ultimately less essentialist than others, adopting a Wittgensteinian conception of the vocabulary of 'creativity,' according to which it is part of a 'language game' involving many overlapping and competing definitions. However, she nonetheless opts for a heuristic process-first definition that excludes AI creativity. For a comparable position, see Nanay (2018, 23).

*Anthropomorphizing AI*. The second denialist approach is related to the first, and they usually go hand in hand. According to this response, ordinary language users mistakenly anthropomorphise AI. From this perspective, people do not *neglect* the IAC; rather, they mistakenly assume it to have been fulfilled. As du Sautoy (2019, 237) puts it: "Often we respond to code that we don't understand by assigning it some sort of [conscious] agency." Or in the words of Runco (2023: 3): "it makes no sense to believe that machines have intentions, but humans may attribute them with intentions" (see also Bender et al. 2021: 616). On this view, people fallaciously adopt what Dennett (1998) calls the "intentional stance" towards AI, where they try to explain and predict the actions of AI systems based on inferred intentions. Otherwise put, people think of AI as 'desiring' to achieve certain ends. So perhaps when people refer to AI as 'creative' they are not consciously or unconsciously overlooking the IAC – instead, they wrongly presume it to have been satisfied. According to this objection, we can once again reject ascriptions of creativity to AI as straightforwardly incorrect.

This argument in part rests upon an empirical claim about how people are perceiving AI when making judgments concerning its creativity. This is amenable to empirical testing, and some relevant experimental work has been conducted on this topic. For example, a survey-based study by Moruzzi (2022) found that people tend to judge both the agency and creativity of artificial agents significantly lower in comparison to humans, and that these two types of judgment (of agency and creativity) tend to be positively correlated. However, the study does not demonstrate that people are consistently ascribing agency to AI systems prior to ascribing creativity thereto. There is nonetheless ample evidence that people anthropomorphise AI (see Li and Suh 2022 for an overview), and multiple studies indicating that people are prone to attribute intentional agency to generative AI (for an overview, see Salles et al. 2020). But the empirical literature ultimately remains divided on this topic, and a recent survey found that the majority of people living in the US do *not* believe AI to be conscious (Dreksler et al. 2025). It therefore remains unclear to what extent people's ascriptions of creativity to AI are conditioned by anthropomorphizing projections of conscious intentional agency.

This second denialist approach is also vulnerable to the objections that we surveyed in response to the first approach: it is heavily prescriptivist, and jars with the significant number of experts who ascribe creativity to AI while *explicitly* rejecting the idea that AI is capable of conscious, intentional agency (e.g., Helliwell 2026b; Langland-Hassan 2026; McClelland 2026). This suggests that at least some people are not projecting intentional agency onto AI when classifying it as creative. Langland-Hassan (2026) and Boden (2014) have forcefully argued that most artists are not usually pursuing conscious intentions when exercising their creativity. One could therefore reply to the denialists that it is potentially they who are mistakenly projecting intentional agency – that is, onto *human* creativity.

*Derivative Intentions.* The third denialist response involves arguing that intentional agency *is* causally driving AI's creative outputs – however, this agency is derived from the programmers who write the algorithms that produce these outputs, along with those who curate the data sets on which such systems are trained, and furthermore, with those who write, or "engineer" the relevant prompts (see e.g., Millière 2022). According to this response, creativity is being incorrectly ascribed – people should be attributing it to the network of human agents who design and prompt an AI system, rather than to the AI system itself. This objection aligns with the view held by some moral philosophers that the actions of AI should invariably be considered products of human agency (e.g. Popa 2021).

The problem with this objection is that none of these human agents can accurately envision what the AI system is going to produce. To begin with the prompters: the prompting usually involved in an AI-generated creative output is minimal and certainly does not amount to genuine creative agency (Nannicelli 2025), especially given that many models (e.g., Leonardo) have built-in prompt-refining LLMs that revise the user's prompts and as such heavily dilute their agency. Regarding the programmers, generative AI systems often produce creative results that the programmers in no way foresaw, and hence they cannot be taken to have intended these outputs (Boden 2014). The case may be different with small-scale generative models developed or curated by the very same agents who prompt the creative output, but in the case of large-scale LLMs and diffusion models, such as ChatGPT or Gemini, creative intention cannot be ascribed to prompters, programmers, or model trainers. One response to this is to argue that the intentional agency involved in AI creativity is diffuse, being spread across multiple human actors, interacting with human-dependent computational systems (see e.g., Zeilinger 2021; Tigre Moura 2023). However, once again, this clashes with the increasingly widespread intuition that the driving source of creativity lies in the generative AI systems themselves, and not with programmers, curators, or prompters (Nannicelli 2025).

These are the main candidate solutions for resolving the IAC predicament by *denying* the possibility of AI creativity, and as we can see, while none can be ruled out as conclusively incorrect, none of them are satisfactory. We should therefore now examine whether accommodationist responses fare any better.

4.2. Accommodationist Responses

*Product-First.* The most straightforward accommodationist response is to adopt what is often called a 'product-first' conception of creativity. This approach is usually presented in opposition to the process-first approach (Moruzzi 2025) and involves determining whether agents are creative solely based on whether they generate outputs – i.e., *products* – that are considered novel and valuable, regardless of the causal processes by which these outputs were generated. Theresa Amabile, one of the foremost proponents of the product-first approach, glosses it as follows: "judges of creativity shouldn't need to know where a product came from. If they saw it as novel relative to other products within its domain, and appropriate for some task or problem, then we would have to accept their judgments as such" (Amabile 2020). According to this view, the IAC can be disregarded, since that would constitute part of the generative process, extrinsic to the product itself. Although this is a coherent solution to the IAC predicament, it fails to gel with the dominant belief among experts and laypeople that processes do indeed matter. Gaut's toy example of knocking over the set of paints has traction here. We do not tend to consider an agent that produces something original and novel purely by accident to be engaging in creativity. So clearly, at least some of the time, generative processes condition our ascriptions of creativity, especially when we are ascribing creativity to agents (which is our focus is our focus in this paper). Consequently, while the product-first view constitutes a valuable response, it is, like the denialist approach, too sharply at odds with current lay and expert classificatory practices, and is therefore too forcefully prescriptive to represent a convincing solution.

*AI Satisfies the IAC.* The next accommodationist approach involves arguing that it is perfectly reasonable to ascribe intentional agency to generative AI, and hence that AI can in fact satisfy the IAC. This response maintains that AI might well be conscious and correspondingly quite capable of satisfying the IAC. On this view, people are not *projecting* consciousness onto generative AI systems, they are correctly identifying it. Probably due to the radical nature of the argument, nobody has yet tried to defend it in a comprehensive manner. Several theorists have, however, floated this idea as a hypothetical possibility. Langland-Hassan (2026: 550), for example, casts doubt on denialist approaches to creativity by asking "whether (and how) we know that generative AI tools are not conscious." Others remain open to the possibility that AI consciousness may emerge in the future, which would make AI creativity perfectly possible even if we cleave to the IAC (Boden 2014; Colton 2019; du Sautoy 2019). However, this would still not resolve the IAC predicament, since many theorists who predicate creativity to AI do so while explicitly assuming AI to *not* be conscious. And as we have also seen, there is evidence that most non-experts likewise do not take AI to be conscious (Dreksler et al. 2025). As a result, even if AI happened to be conscious, and to objectively satisfy the IAC, this would still not shed much light on the concept of creativity that people are operating with – it would merely imply that AI can be considered objectively creative according to one partisan conception of creativity, and that general usage fortuitously happens to be coinciding with this conception.

*Reconceiving Intentional Agency.* The third accommodationist response is to argue that one does not need to suppose that a system is conscious in order to ascribe intentional agency to it. This idea has roots in Dennett's claim that it is perfectly legitimate to project intentional agency onto computational systems by adopting an "intentional stance" towards them – namely, because we are similarly projecting intentional agency onto humans when we adopt this stance towards them (Dennett 1998). Whenever we adopt the intentional stance, we do so without genuine knowledge of any underlying mental states, and the causal impact these might have had on observable behaviour. What nonetheless warrants our ascriptions of conscious intention to particular types of agent is the predictive power of such ascriptions. In line with Dennett's thesis, others have similarly argued that AI systems do not need to possess consciousness in order for us to attribute intentional agency to them – they need only be constitutionally *goal-oriented* (Dung 2025; Ward et al. 2024; Ventura 2019; Guckelsberger et al. 2017).[8] What is more, several theorists of creativity have pointed out that human creativity often does not involve conscious agency – that artists often produce their works without any clear idea of what it is they are creating, and are unable to explain *post hoc* how they went about creating the output (Langland-Hassan 2026; Boden 2014).

McClelland (2026: 7) identifies a problematic feature of this dispute. He quite rightly argues that "it costs the advocate of artificial creativity nothing to take a stand on this debate and maintain that consciousness is not a pre-requisite of agency," though he points out that this merely leads to a "dialectical stalemate". However, if, as we claim below, the motivations for the IAC can be altogether accounted for without appealing to intentional agency as a necessary condition of creativity, then we can sidestep this impasse concerning the nature of intentional agency. Accordingly, we will remain agnostic as to whether we need to reconceive intentional agency in the manner suggested by many theorists.

[8] On the topic of ascribing agency to AI, see Floridi (2025).

One relevant upshot of these reflections concerning goal-orientedness is that it neutralizes the 'accidental creativity' counterexample. It does this by bringing into relief the fact that non-human systems *can* coherently generate outputs without conscious intentional agency, and yet without this generation being 'accidental' in the manner of the paint set being spilled onto the floor. Nor is the productive process blind in the manner of evolutionary creativity, insofar as goals – understood as program end points – are present within the system before it commences the generative process.

*The Tuning Model.* In response to observations along these lines, Langland-Hassan (2026) has argued that we would do better to think of creativity as requiring a form of 'tuning' between the productive system and its outputs.

> It is no accident that when you strum a guitar tuned to an open chord the resulting sounds harmonize. In the same way, a generative DNN [deep neural network] is tuned during its training across millions of parameters. The kind of activation that eventually ramifies through the network is very far from an accident, even if it is not (necessarily) the result of an intention. In the same way, [the musician] Neil Young has been intricately tuned by his past experiences and genetics to be the complex, distinctive songwriter that he is. This tuning can likewise ring out in works that spring from his subconscious without being accidents or flukes, and yet without being intentionally generated. (Langland-Hassan, 2026: 552)

The tuning model of creativity carves out a conceptual space for non-accidental but nonetheless unintended creativity. While we believe that the tuning conception of creativity sets us out on the right path, it has some shortcomings that mean that it does not satisfactorily resolve the IAC predicament. For one, the notion of 'tuning' is too ambiguous and underdeveloped. Langland-Hassan sometimes indicates that this is a kind of 'harmony' – that the system is creative if its outputs are the 'ringing out' of an underlying structure – in the case of the artist, their personality; in the case of a DNN, the underlying generative algorithm, the training set, and the parameter tunings. However, this is not sufficient to ascribe creativity since a single 'ringing out' would not be enough for an external judge to establish that there is indeed tuning present. This could still be a fluke. Or it could be the first instance in a formulaic series of cookie-cutter outputs, that subsequently lack novelty despite expressing an underlying 'tuning'. We would not call a machine creative if it only produced one painting and then broke down, or if it rang-out its inner tuning in the same outputs over and over again. We therefore need to know that an agent is 'tuned' in a fashion that gives them the stable competence to express their tuning repeatedly and in ways that are both novel and valuable, across multiple outputs. Moreover, this cannot be a product of purely randomized production. The monkey at the typewriter that occasionally writes an original and valuable poem is not thereby exhibiting creativity.

Another issue with the tuning model is that it does not sufficiently defuse the 'praise intuition' that was a core justification for the IAC, nor does it explain why we typically take intentional agency to be so important. In Langland-Hassan's words, the praise-like quality of ascriptions of creativity is simply a contingent byproduct of the fact that we have a "natural inclination to praise artists (and other creative individuals) for their works, combined with the reasonable intuition that it makes no sense to praise mere accidents or flukes" (Langland-Hassan 2026: 552). Consequently, he maintains that we can simply drop the idea that the language of creativity *necessarily* expresses praise.[9] However, this too readily dismisses the praise intuition as a contingent byproduct of certain types of human creativity when even ascriptions of creativity to AI seem to have a praise-like quality. As such, while we are sympathetic to Langland-Hassan's tuning solution, it does not as it stands resolve the IAC predicament.

What the tuning model lacks, then, is an account of our intuitions that a) ascriptions of creativity require the agent to be capable of *repeatedly* producing novel and valuable outputs in a non-random manner; and b) intentional agency and praise-giving are integrally important for such ascriptions. In the following section, we use the tuning model as a basis for formulating a new and more robust accommodationist solution to the IAC predicament – one that we believe fills these gaps.

### 4.3. An Alternative Solution: Inferring Creative Ability

The key to our approach lies in Boden's statement that "Creativity is the *ability* to come up with ideas or artefacts that are new, surprising and valuable" (Boden 2004: 1; emphasis added). The fact that a generative agent expresses their tuning in a novel and valuable output does not entail that the agent possesses any creative *ability* – for one, as we saw, this could still be a one-off. There are complex debates regarding what counts as an ability (for an overview, see Jaster

[9] Currie and Turner (2023) likewise seek to jettison the praise intuition.

2020), but for our purposes, we adopt the heuristic definition of an ability as the counterfactual capacity of an agent to engage in a specific type of activity.

The idea of creativity as an ability ties in with accounts of creativity as a competence or virtue (Kieran 2014), or as a skill (Gaut 2009; Helliwell 2026a). That said, there are important differences. Conceiving creativity as a *virtue* is anthropocentric and morally shaded and therefore jars with the fact that people regularly ascribe creativity to non-human agents, as well as to vicious forms of behaviour, such as the design of torture devices (what is typically called 'malevolent' creativity) (see Pearson 2021). Creativity may *usually* be a virtue in the case of humans, but this does not make it *necessarily* so.

Similarly, framing creativity as a *skill* is too focused on the necessity of certain acquisitional processes. Although Helliwell (2026a) has sought to show that AI systems can coherently possess "skills", she follows Gaut in setting the process of *acquisition* as a key criterion of whether an agent possesses a skill, which she describes as "a kind of accomplishment … that can be practiced … [and] that can be learnt." By contrast, thinking of creativity as an *ability* does not require any knowledge of the acquisitional process. This appears to be more warranted given the widespread tendency to ascribe creativity to natural processes and deities (such as the Christian God), even though it seems out of the question to consider either of these as having "acquired" "practiced" or "learnt" to be creative.

Our point is not that the story of acquisition is irrelevant for empirical accounts of how creativity is developed, but just that it misses what actually warrants ascriptions of creativity, which is rather the perception of creative functional organization, i.e., creative *ability*. To judge an output and its source as creative, and not just as a fluke, we have to be able to infer the existence of a creative functional organization inhering within the agent and acting as the principal cause of its novel and valuable output(s). In order for this type of organization to count as an *ability*, however, the agent needs to be capable of manifesting their inner tuning on *multiple* occasions, at least counterfactually, and not as product of merely random generative behaviour. To ascribe creativity to a given agent, we therefore need to be confident that the agent is functionally organized in a way that means that they have already generated at least one novel and valuable output, and are *ceteris paribus* capable of doing so again in the future in a goal-directed (i.e., non-random) fashion.

There are multiple inferential pathways by which we can arrive at the reasonable belief that a system possesses this type of functional organization. This is not a matter of fulfilling necessary criteria; it is a question of garnering sufficient evidence to carry us over the threshold of confidence that warrants our inferring such organization, and this evidence can derive from mixed sources. We do not have space to give a comprehensive account of this inferential procedure, so in what follows we will merely sketch its outlines.

*Repeated performance*. The first and most obvious way that we infer such organization is by observing repeated performance. This is one of the most reliable ways by which we can ascertain whether a single creative output was a one-off fluke or the product of a real ability. And this applies in the case of humans and non-human agents alike. If a black box could repeatedly produce creative masterpieces, but without being able to communicate intention or understanding to humans, we would still call the black box creative, simply on the grounds of its consistency. Similarly, people call evolution creative because of its well-proven ability to generate novel and original biological phenomena.

*Perceived non-random contingency*. Mere rule-following does not feel like genuine creative ability – namely, because it does not give us the impression that the agent in question could continue to produce novel and valuable outputs, because it is likely to end up repeating itself. But on the other end of the spectrum, totally random generation similarly does not strike us as a sign of creative ability, since it is indistinguishable from the accidental generation of novel and valuable outputs. A creative agent therefore needs to contain some element of controlled unpredictability or *contingency*. There needs to be room for future novelty and the generation of surprise by what we can infer to be non-random processes. And to be sure, as many have noted, this can certainly be the case with DNNs. For one thing, with many generative AI systems, external input from humans is continually modifying the token and parameter weightings, and hence a single prompt is vanishingly unlikely to produce the same output twice. What is more, DNNs are stochastic and parameters such as temperature, random seeds, and 'Top P' are typically set so that the LLM selects less likely (but not random) tokens, which introduces a significant element of unpredictability, or *entropy*, into the system without this amounting to pure randomness. As such, the rule-based, algorithmic organization of generative AI systems is perfectly compatible with a sense that they contain sources of contingency – that is, sufficiently high entropy – and could therefore generate novel and valuable ideas in the future in a controlled fashion.

*Perceived intentions* and *reflective understanding.* Another route by which we can infer creative organization is by checking for an *intention* antecedent to the creative act. If we perceive that a person consciously intends to do something – e.g., because they explicitly state their intentions, or because we have asked them to perform a particular action – before they proceed to successfully do so, this can be taken as a strong indication that the performance resulted from a real ability to realize a causal process that reliably yields the intended activity.

Relatedly, we have further warrant to infer creative ability if the agent can then, *post hoc*, explain how and why they created what they intended to create (see Gaut 2010). In doing so, they exhibit a type of *reflective understanding* that further indicates functional organization, and correspondingly that they did not generate the novel and valuable output in question by accident.

*Design.* Another source of warrant for ascribing an ability to an agentic system is knowledge that it has been *designed* to exercise that ability. Thus, if we know that a system has been engineered to produce novel and valuable outputs, then even if it only does so a small number of times, we can be relatively confident that it nonetheless possesses an underlying organization 'geared' to do so, and that it would be able to do so in the future. This does not, however, mean that the resulting creativity is entirely ascribable to the intention of the designer. As we have seen in the case of AI, programmers usually do not know what kinds of creative products their algorithms and training sets are going to generate, and so while knowledge of design may give us warrant to infer creative ability in the agentic system, it does not warrant our attributing the resulting creativity to the intentions of the designer of that system.

*Responsiveness.* Another means by which we can infer creative functional organization is if we see that the agentic system is "responsive" to its environment, that it can adapt its outputs based on registered obstacles to its goals (see Ward et al. 2024). If we see that an agent adapts in the face of obstacles in the process of generating a novel and valuable output, we can be confident that it is internally organized in an end-oriented manner directed toward the production of such outputs. Currie and Turner (2023: 122), following Gaut, refer to this as "sensitivity", and argue that evolutionary systems exhibit precisely this type of responsiveness when they overcome adaptive hurdles. Much the same applies to generative AI. If we interrupt an LLM composing a creative story with a prompt telling the LLM that we do not find the output creative, the LLM can stop and adapt its strategy to produce something that we are more likely to consider creative. Similarly, a generative adversarial network evaluates its outputs, refining and adapting them if they do not fit the initial prompt.

Thus, *repeated performance, perceived non-random contingency*, *perceived intention, reflective understanding, design,* and *responsiveness* can all act as signals of an underlying functional organization or *ability* – in this case, to generate original and valuable outputs. And to reiterate, none of these signals are *necessary*. In different cases, they can contribute in differing degrees to warranting the attribution of creative ability to a given agent.

We should respond to one objection to our suggestion that creativity consists in being counterfactually capable of repeatedly generating novel and valuable outputs: What about stellar one-offs? Take Emily Brontë's *Wuthering Heights*. But in such a case, which is certainly not a fluke, we can see that the creative process took time, and was composed of innumerable smaller creative feats, that Brontë responded to obstacles, and we know from biographical documents that Brontë carefully planned the novel, or in other words, *intended* it. From this information, we are warranted to suppose that had she lived longer, she almost certainly *would have* been able to produce other novel and valuable works, even if she had chosen not to. So-called one-hit wonders can, therefore, still give us grounds to infer creative ability.

One advantage of this view is that it fits with the widespread intuition that natural systems, as well as generative AI, can be considered creative, but it also explains why we have a weaker sense of their creativity as compared with creative *humans* – because we can get an impressively fine-grained view of how humans have generated their works. Consequently, we can infer with far greater confidence when humans are functionally organized in a way that enables them to generate novel and valuable outputs in a non-random fashion. It also accounts for the intuition that intention is of prime importance when ascribing creativity, but it does so without positing intentional agency as a necessary condition of creativity.

A further advantage of this theoretical view is that we do not need to abandon the idea underlying the intuition that ascriptions of creativity are approbative. However, we do need to reconceive them as a more general expression of *approval* – we approve of creative functional organization because it tends to provide us with new and valuable products (though dark creativity is an important exception to this rule). In the case of humans, this expression of approval would usually take the form of praise, but in the case of generative AI, it would simply remain a basic expression of *approval* without any attribution of desert. This naturally does not mean that we need to approve of generative AI *in the round*, just that on some level we approve its ability to generate novel and valuable outputs when we label this generation 'creative.' This enables us to hold onto our intuition that the language of creativity tends to be positively valanced.

On our accommodationist model, we can therefore call an agent creative when we have grounds to confidently infer that the efficient cause of their original and valuable outputs is their creative *ability*. Our *creative abilities* solution makes good the deficiencies of the tuning model insofar it accounts for our intuitions that a) ascriptions of creativity require the agent to be capable of *repeatedly* producing novel and valuable outputs in a non-random manner; and b) intentional agency and praise-giving are integrally important for such ascriptions. We believe this account of how judgments of creativity are made is superior to existing accounts because it offers a plausible explanation of why people are inclined to ascribe creativity to generative AI systems, and it does so without framing these ascriptions as erroneous. Our proposed solution therefore resolves the IAC predicament posed in Section 2: it explains how creativity can be ascribed to AI while also accounting for the intuitions that originally motivated the IAC and denied the coherence of AI creativity.

## Conclusion

In this paper we have pushed back against the idea that creativity requires conscious intentional agency (the IAC). We saw that the IAC is motivated by intuitions about the language of creativity that, since the advance and proliferation of generative AI, have come under heavy strain. We collated and critically analysed the various candidate responses to this predicament. We first surveyed denialist responses, which double down on the IAC and insist that ascribing creativity to AI is simply incorrect. Although we have not sought to disprove this essentialist position, we have argued that it is excessively prescriptive, and that many of the intuitions on which it is based are increasingly open to doubt. We then examined accommodationist solutions, finding them to offer a more promising way forward, insofar as they keep theoretical discussions of creativity aligned with natural language use. However, we found that the most plausible accommodationist proposals were nonetheless flawed. To resolve these shortcomings, we formulated a solution that frames robust inferences of *creative ability* as the key source of warrant for ascriptions of creativity.

Further empirical research is certainly required, however, to assess the extent to which people are anthropomorphizing AI when ascribing creativity thereto. There is good reason to think this is occurring to some extent, even if it does not suffice to explain the increasing acceptance of AI as genuinely creative. This is especially pressing given that the anthropomorphism of AI may inadvertently foster the recognition of AI as a moral agent (Salles et al. 2020). Such recognition would have serious consequences for societies and should only be granted on well-considered grounds. Any increase in implicit recognition therefore ought to be closely monitored, especially given that any such shift could be driven by marketing hype.

Our corpus analyses also point to the need for further linguistic research into how the proliferation and promotion of AI is impacting natural language use, especially with regard to the language of creativity. The two corpus analyses that we have presented in this study are adduced as suggestive – and not conclusive – evidence of a broader trend towards ascribing creativity to AI that we see in the recent theoretical literature on creativity. However, further work is required to quantify and explain these changes, and to establish whether they are indicative of a general sea change in people's considered attitudes towards creativity. It would be particularly interesting to perform a cross-cultural analysis, including non-anglophone corpuses, to determine whether the same trends can be identified interculturally.

Whichever direction such future research into AI creativity takes, we believe that it is vital to adopt an ecumenical definition of creativity that does not exclude artificial intelligence or demote its productive abilities to the level of mere pseudo-creativity. We need to take seriously the possibility that advances in artificial intelligence are modulating our linguistic, philosophical, and moral intuitions. The accommodationist approach that we have defended here enables us to remain open to this possibility without making the speculative leap of ascribing consciousness, or other human cognitive faculties to artificial intelligence systems.

## References

Amabile, Teresa. 2020. "Creativity, Artificial Intelligence, and a World of Surprises." *Academy of Management Discoveries* 6 (3): 351–54.

Arnheim, Rudolf. 2001. "What it Means to be Creative." *British Journal of Aesthetics* 41 (1): 24–5.

Bender, Emily M., Timnit Gebru, Angelina McMillan-Major, and Shmargaret Shmitchell. 2021. "On the Dangers of Stochastic Parrots: Can Language Models Be Too Big?" In *Proceedings of the 2021 ACM Conference on Fairness, Accountability, and Transparency*, 610–23. New York: Association for Computing Machinery.

Boden, Margaret. 1998. "Creativity and Artificial Intelligence." *Artificial Intelligence* 103 (1–2): 347–56.

Boden, Margaret. 2004. *The Creative Mind: Myths and Mechanisms.* 2nd ed. London: Routledge.

Boden, Margaret. 2007. "Authenticity and Computer Art." *Digital Creativity*, 18 (1): 3–10.

Boden, Margaret. 2014. "Creativity and Artificial Intelligence: A Contradiction in Terms?" In *The Philosophy of Creativity*, edited by Elliot Samuel Paul and Scott Barry Kaufman, 224 –244. New York: Oxford University Press.

Boden, Margaret. 2019. "Art Appreciation and Creative Skills." In *From Fingers to Digits: An Artificial Aesthetic*, edited by Margaret Boden and Ernest Edmonds, 91–106. Cambridge, MA: The MIT Press.

Brainard, L. 2025. "The Curious Case of Uncurious Creation." *Inquiry*, 68(4), 1133–1163.

Colton, Simon. 2019. "From Computational Creativity to Creative AI and Back Again." *Interalia Magazine*. September. https://www.interaliamag.org/articles/simon-colton/.

Currie, Adrian, and Derek Turner. 2023. "Creativity Without Agency: Evolutionary Flair and Aesthetic Engagement." *Ergo* 10 (4). https://doi.org/10.3998/ergo.4633.

Davies, Mark. 2016–. Corpus of News on the Web (NOW). Available online at https://www.english-corpora.org/now/. Accessed 3 Mar 2026.

Dennett, Daniel. 1998. *The Intentional Stance*. Cambridge (MA): MIT Press.

Dreksler, Noemi, Lucius Caviola, David Chalmers, Carter Allen, Alex Rand, Joshua Lewis, Philip Waggoner, Kate Mays, and Jeff Sebo. 2025. "Subjective Experience in AI Systems: What Do AI Researchers and the Public Believe?" https://doi.org/10.48550/arXiv.2506.11945.

Dung, Leonard. 2025. "Understanding Artificial Agency." *Philosophical Quarterly* 75 (2): 450-472.

Floridi, L. 2025. "AI as Agency without Intelligence." *Philos. Technol.* 38, 30. https://doi.org/10.1007/s13347-025-00858-9.

Gaut, Berys. 2003. "Creativity and Imagination". In *The Creation of Art*, edited by B. Gaut and P. Livingston, 148–73. Cambridge: Cambridge University Press.

Gaut, Berys. 2009. "Creativity and Skill." In *The Idea of Creativity*, edited by Michael Krausz, Denis Dutton, and Karen Bardsley, 83–103. Leiden: Brill.

Gaut, Berys. 2010. "The Philosophy of Creativity." *Philosophy Compass* 5 (12): 1034–46.

Gaut, Berys. 2018. "The Value of Creativity." In *Creativity and Philosophy*, edited by Matthew Kieran and Berys Gaut, 124–139. London: Routledge.

Google. 2024. "Google Ngram Viewer". Google Inc. https://books.google.com/ngrams/. Accessed 3 March 2026.

Green, D. *Serendipity Machine*. 2004. Crows Nest, NSW: Allen & Unwin.

Guckelsberger, C., C. Salge, and S. Colton. 2017. "Addressing the 'Why?' in Computational Creativity: A Non-anthropocentric, Minimal Model of Intentional Creative Agency." In *Proceedings of the Eighth International Conference on Computational Creativity,* edited by A. Goel, A. Jordanous, and A. Pease. Atlanta: Association for Computational Creativity.

Guzik, E. E., C. Byrge, and C. Gilde. 2023. "The Originality of Machines: AI takes the Torrance Test." *Journal of Creativity* 33 (3). https://doi.org/10.1016/j.yjoc.2023.100065.

Haase, J., and P. H. P. Hanel. 2023. "Artificial Muses: Generative Artificial Intelligence Chatbots have Risen to Human-Level Creativity." *Journal of Creativity* 33 (3). https://doi.org/10.1016/j.yjoc.2023.100066.

Halina, M. 2021. “Insightful Artificial Intelligence.” *Mind & Language* 36 (2): 315–329.

Helliwell, Alice C. 2026a. “Can AI Create with Flair? Artistic Creativity and Artificial Intelligence.” *The British Journal of Aesthetics*. https://doi.org/10.1093/aesthj/ayaf057.

Helliwell, Alice C. 2026b. “Creativity, Agency, and AI.” In *Philosophy of Artificial Intelligence*, edited by Müller, V. C., Dung, L., Löhr, G., Rumana, A. Cham: Springer. https://doi.org/10.1007/978-3-032-10073-3_13 .

Hoel, Erik. 2022. “AI-Art Isn’t Art.” *The Intrinsic Perspective,* May 18. https://www.theintrinsicperspective.com/p/ai-art-isnt-art.

Hubert, K., K. Awa, and D. Zabelina. 2023. “Artificial Intelligence is More Creative Than Humans.” *PsyArXiv*, September 18, 2023. doi:10.31234/osf.io/6py2m.

Hutson, Matthew. 2023. “Hypotheses Devised by AI Could Find ‘Blind Spots’ in Research.” *Nature.* November 17. https://doi.org/10.1038/d41586-023-03596-0.

Ivcevic, Z., and M. Grandinetti. 2024. “Artificial Intelligence as a Tool for Creativity.” *Journal of Creativity,* 34 (2). https://doi.org/10.1016/j.yjoc.2024.100079.

Jaster, Romy. 2020. *Agents’ Abilities*. Berlin: DeGruyter.

Jordanous, A., and Keller, B. 2016. “Modelling Creativity: Identifying Key Components through a Corpus-Based Approach.” *PloS One*. 11(10). https://doi.org/10.1371/journal.pone.0162959.

Kelly, Sean. 2019. “A Philosopher Argues That an AI Can’t Be an Artist.” *MIT Technology Review.* February 21. https://www.technologyreview.com/s/612913/a-philosopher-argues-that-an-ai-can-never-be-an-artist/.

Kieran, Matthew. 2014. “Creativity as a Virtue of Character.” In *The Philosophy of Creativity*, edited by Elliot Samuel Paul and Scott Barry Kaufman, 125–144. New York: Oxford University Press.

Kind, Amy. 2022. *Imagination and Creative Thinking*. Cambridge: Cambridge University Press.

Langland-Hassan, Peter. 2026. “Imagination, Creativity, and Artificial Intelligence.” In *The Oxford Handbook of Philosophy of Imagination and Creativity*, edited by Amy Kind and Julia Langkau, 543–557. Oxford: Oxford University Press.

Li, M. and A. Suh. 2022. “Anthropomorphism in AI-enabled technology: A literature review.” *Electron Markets* 32, 2245–2275.

Livingston, Paisley. 2018. “Explicating Creativity.” In *Creativity and Philosophy*, edited by Matthew Kieran and Berys Gaut, 108–23. London: Routledge.

Lohr, Steve. 2017. “A.I. Will Transform the Economy. But How Much, and How Soon?” *The New York Times*, November 30. https://www.nytimes.com/2017/11/30/technology/ai-will-transform-the-economy-but-how-much-and-how-soon.html.

Marchant, Jo. 2025. "Can AI be Truly Creative?" 2025. *Nature* 647: 24–26.

McClelland, T. 2026. "Does artificial creativity require artificial consciousness?" *AI & Society*. https://doi.org/10.1007/s00146-026-02887-0.

Metz, Cade. 2016. "Google's AI Wins Pivotal Second Game in Match with Go Grandmaster." *Wired*, March 10. https://www.wired.com/2016/03/googles-ai-wins-pivotal-game-two-match-go-grandmaster/.

Michel, J. B., Y. K. Shen, A. P. Aiden, A. Veres, M. K. Gray, Google Books Team, J. P. Pickett, D. Hoiberg, D. Clancy, P. Norvig, J. Orwant, S. Pinker, M. A. Nowak, and E. L. Aiden. 2011. "Quantitative Analysis of Culture Using Millions of Digitized Books." *Science* 331 (6014): 176–182.

Millière, Raphaël. 2022. "AI Art Is Challenging the Boundaries of Curation." *Wired,* July 17, 2022. https://www.wired.com/story/dalle-art-curation-artificial-intelligence/.

Mollick, Ethan. 2023. "Automating Creativity." *One Useful Thing*. August 13. https://www.oneusefulthing.org/p/automating-creativity.

Moruzzi, Caterina. 2022. "Creative Agents: Rethinking Agency and Creativity in Human and Artificial Systems." *Journal of Aesthetics and Phenomenology* 9 (2): 245–268.

Moruzzi, Caterina. 2025. "Artificial Intelligence and Creativity." *Philosophy Compass* 20 (3). https://doi.org/10.1111/phc3.70030.

Nanay, Bence. 2018. "An Experiential Account of Creativity." In *The Philosophy of Creativity*, edited by Elliot Samuel Paul and Scott Barry Kaufman, 17–38. New York: Oxford University Press.

Nannicelli, Ted. 2025. "Mass AI-art: A Moderately Skeptical Perspective." *The Journal of Aesthetics and Art Criticism* 83 (4): 343–357.

Paul, Elliot Samuel, and Dustin Stokes. 2018. "Attributing Creativity." In *Creativity and Philosophy*, edited by Matthew Kieran and Berys Gaut, 193–209. London: Routledge.

Pearson, James S. 2021. "The Value of Malevolent Creativity." *Journal of Value Enquiry* 55, 127–144. https://doi.org/10.1007/s10790-020-09741-6.

Popa, E. 2021. "Human Goals Are Constitutive of Agency in Artificial Intelligence." *Philosophy and Technology* 34, 1731–1750 (2021). https://doi.org/10.1007/s13347-021-00483-2.

Runco, Mark A. 2023. "AI Can Only Produce Artificial Creativity." *Journal of Creativity* 33 (3). https://doi.org/10.1016/j.yjoc.2023.100063.

Runco, Mark, and Garrett Jaeger. 2012. "The Standard Definition of Creativity." *Creativity Research Journal* 24 (1): 92–96. https://doi.org/10.1080/10400419.2012.650092.

Salles, A., K. Evers, and M. Farisco. 2020. "Anthropomorphism in AI." *AJOB Neuroscience*, 11 (2): 88–95. https://doi.org/10.1080/21507740.2020.1740350.

Stokes, Dustin. 2011. "Minimally Creative Thought." *Metaphilosophy* 42: 658–81.

Stokes, Dustin. 2014. "The Role of Imagination in Creativity." In *The Philosophy of Creativity*, edited by Elliot S. Paul and Scott Barry Kaufman, 157–84. Oxford: Clarendon Press.

Stokes, Dustin, and Elliot Paul. 2021. "Computer Creativity is a Matter of Agency." *IAI News*, November 11. https://iai.tv/articles/ai-will-never-rival-picasso-auid-1971.

du Sautoy, Marcus. 2019. *The Creativity Code*. Cambridge (MA): Belknap.

Tigre Moura, Francisco. 2023. "Artificial intelligence, creativity, and intentionality." *The Journal of Creative Behavior* 57 (3): 336-338.

Ventura, Dan. 2019. "Autonomous Intentionality in Computationally Creative Systems." In *Computational Creativity: The Philosophy and Engineering of Autonomously Creative Systems*, 49–69. Cham: Springer International Publishing.

Ward, Francis Rhys, Matt MacDermott, Francesco Belardinelli, Francesca Toni, and Tom Everitt. 2024. "The Reasons that Agents Act: Intention and Instrumental Goals." In *Proc. of the 23rd International Conference on Autonomous Agents and Multiagent Systems (AAMAS)*. https://arxiv.org/pdf/2402.07221

Zeilinger, Martin. 2021. *Tactical Entanglements: AI Art, Creative Agency, and the Limits of Intellectual Property*. Meson Press. Lüneberg.